%% file: sample_FG2021.tex
\documentclass[a4paper, 10pt, conference]{ieeeconf}      
\usepackage{todonotes}
\usepackage{amsmath}
\usepackage{hhline}
\usepackage{colortbl}
\usepackage{array}
\usepackage{amssymb}
\usepackage{multirow}
\usepackage[affil-it]{authblk}

\newcommand{\specialcell}[2][c]{%
\begin{tabular}[#1]{@{}c@{}}#2\end{tabular}}

\IEEEoverridecommandlockouts                              
\title{\LARGE \bf
Domain Generalisation for Apparent Emotional Facial Expression Recognition across Age-Groups
}

\author[1,*]{Rafael Poyiadzi\thanks{${}^{*}$\textrm{Work done while an intern at Facebook AI Applied Research, UK.}}}
\author[2]{Jie Shen}
\author[2]{Stavros Petridis}
\author[3]{Yujiang Wang}
\author[2,3]{Maja Pantic}

\affil[1]{University of Bristol, UK}
\affil[2]{Facebook AI Applied Research, UK} 
\affil[3]{Imperial College London, UK} 

\begin{document}


\maketitle

\begin{abstract}
Apparent emotional facial expression recognition has attracted
a lot of research attention recently. However, the majority of  approaches ignore age differences and train a generic model for all ages. In this work, we study the effect of using different age-groups for training apparent emotional facial expression recognition models. To this end, we study Domain Generalisation in the context of apparent emotional facial expression recognition from facial imagery across different age groups. We first compare several domain generalisation algorithms on the basis of out-of-domain-generalisation, and observe that the Class-Conditional Domain-Adversarial Neural Networks (CDANN) algorithm has the best performance. We then study the effect of variety and number of age-groups used during training on generalisation to unseen age-groups and observe that an increase in the number of training age-groups tends to increase the apparent emotional facial expression recognition performance on unseen age-groups. We also show that exclusion of an age-group during training tends to affect more the performance of the neighbouring age groups.
\end{abstract}

\input{sections/introduction}
\input{sections/methods}

\input{sections/results}

\input{sections/conclusion}

\newpage

{\small
\bibliographystyle{ieee}
\bibliography{bibliography}
}

\end{document}

%% file: sections/introduction.tex
\section{INTRODUCTION}
Apparent emotional facial expression recognition from imagery is an important problem in computer vision with applications in understanding human behaviour and in human-computer interaction. One challenge in automatic apparent emotional facial expression recognition is aging, as it affects facial features, such as wrinkles and folds and facial muscles' elasticity \cite{suja2018pose,nora2019thesis}. 
One of the first works to study the effect of human aging on apparent emotional facial expression recognition is \cite{guo2013facial}. They show that apparent emotional facial expression recognition is influenced by aging, with experiments on the Lifespan \cite{lifespandatabase} and FACES \cite{facesdatabase} databases. They show that expressions exhibited by older people are different from expressions exhibited by younger people, and that they tend to be less exaggerated \cite{nora2019thesis}. It was also found that apparent emotional facial expressions of older people were more difficult to recognise, due to different face-features \cite{AndreaCaroppo:1127}.

There are several works for apparent emotional facial expression recognition  \cite{toisoul2021estimation} but they do not take into account differences in age. They build a generic model for all ages. In this work we want to investigate the effect of using different age-groups during training on the performance of apparent emotional facial expression recognition algorithms.

Machine Learning systems are expected to show good performance when the evaluation domain is similar to the one(s) used for training. When this is not the case, out-of-distribution generalisation, i.e. training and testing apparent emotional facial expression recognition models on different age-groups, is often poor \cite{gulrajani2020search}. In \textit{Domain Generalisation} (DG) the aim is to devise algorithms that perform well in \textit{unseen} domains. These algorithms are often exposed to a variety of diverse domains during training, with the goal of extracting \textit{invariances} from the training-domains, that apply to test-domains \cite{gulrajani2020search}. 

In this work we study DG in the context of apparent emotional facial expression recognition across age-groups. We carry out experiments on AffectNet \cite{mollahosseini2017affectnet} and use EmoFAN \cite{toisoul2021estimation} as out backbone network (Both are described in Section \ref{section:methods}). Our contributions and findings are as follows:
\begin{itemize}
    \item We evaluate domain generalisation algorithms on the basis of out-of-domain-generalisation and show that CDANN \cite{Li_2018_ECCV} performs the best.
    \item We compare out-of-domain-generalisation for training with three age-groups (domains) and with four age-groups, and our results provide evidence that an increase in the number of training domains does indeed improve apparent emotional facial expression recognition performance on unseen domains.
    \item By comparing different configurations of the training age-groups we show that exclusion of an age-group tends to affect more the apparent emotional facial expression recognition performance of the neighbouring age-groups.
\end{itemize}

%% file: sections/methods.tex
\section{METHODS}

\begin{figure}[!t]
    \includegraphics[scale=0.50]{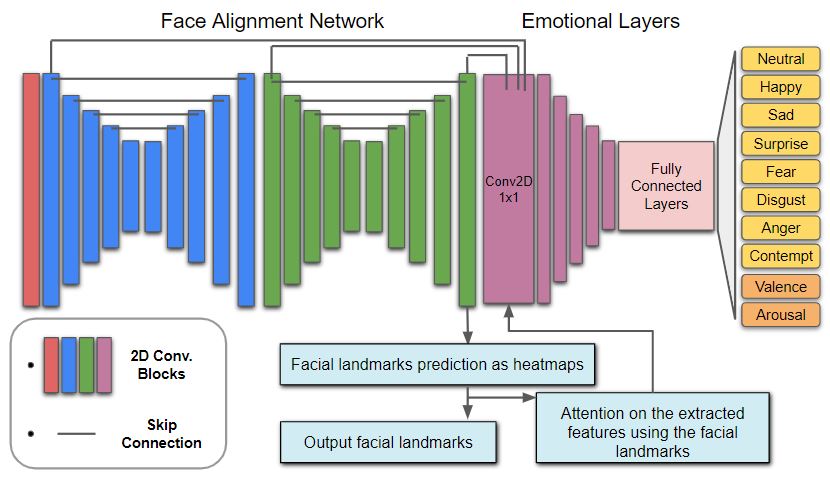}
    \caption{The EmoFAN architecture proposed in \cite{toisoul2021estimation} builds on top of the Face Alignment Network \cite{bulat2017far} to predict jointly discrete apparent emotional classes, continuous affect dimensions and fiducial landmarks on the face. Image recreated from \cite{toisoul2021estimation}.}
    \label{fig:emofan}
\end{figure}

\label{section:methods}
In this section we first present the architecture of our backbone network, then describe the dataset used, and lastly describe the domain generalisation algorithms used.

\subsection{EmoFAN}
The EmoFAN architecture proposed in \cite{toisoul2021estimation} builds on top of the Face Alignment Network \cite{bulat2017far} to predict jointly discrete apparent emotional classes,
continuous affect dimensions and fiducial landmarks on the face. They achieve state-of-the-art performance for apparent emotional facial expression recognition from facial imagery, using both discrete apparent emotion labels and continuous levels of arousal and valence. The network architecture is described in Figure~\ref{fig:emofan}. Low-level description of the architecture can be found in Supplementary Material of \cite{toisoul2021estimation}.

\subsection{Dataset}
AffectNet \cite{mollahosseini2017affectnet} is a large database of apparent emotional facial expressions in-the-wild. The database is annotated with respect to both categorical apparent emotions (\textit{Neutral}, \textit{Happy}, \textit{Sad}, \textit{Surprise}, \textit{Fear}, \textit{Disgust}, \textit{Anger} and \textit{Contempt}) and continuous affect dimensions: \textit{Valence} -how positive the apparent facial emotional display is- and \textit{Arousal} -how calming or exciting the apparent facial emotional display looks \cite{toisoul2021estimation}. From the database we only keep images that: (1) were manually labelled, and that (2) contain a face. For the validation and test splits we use the cleaned versions that were introduced in \cite{toisoul2021estimation}. The apparent age of the subjects in the image was obtained using the approach proposed in \cite{rothe2018deep}.


\subsection{Domain Generalisation}
We introduce the following notation: Let $(x_i^0, y_i^0)$ denote the $i$-th image from domain $[0]$, $\phi_{\theta}(x_i^0)$ denote the output of the \textit{feature extractor} (parametrised with $\theta$) for this image, and let $c_{\theta'}\left(\phi_{\theta}(x_i^0)\right)$ denote the final output of EmoFAN, where $c_{\theta'}$ denotes the \textit{classifier} part of the network (parametrised by $\theta'$. When clear from context we will write $c$ for $c_{\theta'}$ and $\phi$ for $\phi_{\theta}$. In our setup, the fully connect layers of EmoFAN will be referred to as the \textit{classifier} (See pink block in Fig.\ref{fig:emofan}), and everything before it as the \textit{feature extractor}.

In this work we consider five different DG methods:
\begin{itemize}
    \item Empirical Risk Minimisation (\textbf{ERM}) minimises the cumulative loss across domains. The problem is treated as a usual supervised learning task.
    \item Inter-domain Mixup (\textbf{Mixup}) (\cite{wang2020heterogeneous}, \cite{yan2020improve}): The model is trained on convex combinations of randomly chosen image-label pairs coming from different domains. The network is trained using augmented samples of the form:
        \begin{equation*}
            \left(\lambda\phi(x^0) + (1-\lambda)\phi(x^1),~~\lambda y^0 + (1-\lambda) y^1\right)
        \end{equation*}
    where we let $x^0$ and $x^1$ denote batches of data coming from domains $[0]$ and $[1]$ and $\lambda \in (0, 1)$ and is drawn from a Beta distribution.
    
    \item Maximum Mean Discrepancy (\textbf{MMD}): tries to minimise the MMD \cite{gretton2012kernel} across different domains \cite{8578664}. A regularisation term is added to the ERM training as follows:
        \begin{equation*}
            MMD\left(\phi_{\theta}(x^0),~\phi_{\theta}(x^1)\right)
        \end{equation*}
    This regularisation only impacts directly the \textit{feature extractor} part of the network, by trying to minimise the MMD between the distributions of the features of each domain.
    \item Conditional Domain Adversarial Neural Networks (\textbf{CDANN}) \cite{Li_2018_ECCV}: introduces an adversary to match the class-conditional feature distributions. The goal is to make it hard for the per-class conditional feature distributions for every domain to be differentiated by an adversary. For example, make the distributions 
    \begin{equation*}
        \phi(x^0)~|~y^0 = [angry]~~\&~~\phi(x^1)~|~y^1 = [angry]
    \end{equation*}difficult to be distinguished by an adversary.
    \item Meta-Learning for Domain Generalisation (\textbf{MLDG}) \cite{AAAI1816067}: utilises (Model-agnostic Meta-Learning) MAML \cite{pmlr-v70-finn17a} to meta-learn to generalise to unseen domains. 
\end{itemize}

For in-depth comparison consult the cited papers and \cite{gulrajani2020search}\footnote{The implementations of DG algorithms are heavily based on code associated with \cite{gulrajani2020search}.}.

\subsection{Evaluation Metrics}
We use the same loss function as the one proposed in \cite{toisoul2021estimation}. It has four terms: (1) \textit{Cross-Entropy} ($\mathcal{L}_{CE}$) for discrete apparent emotional facial expression recognition, (2) \textit{Mean Squared Error} ($\mathcal{L}_{MSE}$), (3) \textit{Pearson Correlation Coefficient} ($\mathcal{L}_{PCC}$), and (4) \textit{Concordance Correlation Coefficient} ($\mathcal{L}_{CCC}$). MSE, PCC and CCC are computed for both valence and arousal. The loss function is also regularised with shake-shake regularisation coefficients (See $\alpha, \beta \& \gamma$ in Eq.\ref{eq:loss_fn}) that are each sampled from a uniform distribution on $(0, 1)$ on every iteration step \cite{Gastaldi17ShakeShake}.
\begin{align}
    \label{eq:loss_fn}
    &\mathcal{L}(\boldsymbol{y}, \hat{\boldsymbol{y}}) = \mathcal{L}_{CE}(\boldsymbol{y}_{D}, \hat{\boldsymbol{y}}_{D}) + \frac{\alpha}{\alpha+\beta+\gamma}\mathcal{L}_{MSE}(\boldsymbol{y}_{C}, \hat{\boldsymbol{y}}_{C})\nonumber\\
    &+ \frac{\beta}{\alpha+\beta+\gamma}\mathcal{L}_{PCC}(\boldsymbol{y}_{C}, \hat{\boldsymbol{y}}_{C}) + \frac{\gamma}{\alpha+\beta+\gamma}\mathcal{L}_{CCC}(\boldsymbol{y}_{C}, \hat{\boldsymbol{y}}_{C})
\end{align}

We use the following notation: $\boldsymbol{y}$ indicates the set of labels for all subjects, and $\boldsymbol{y}_{D}$ and $\boldsymbol{y}_C$ indicates the discrete and continuous counterparts (referring to valence and arousal) of $\boldsymbol{y}$ respectively. In all cases the \textit{hat} ($\hat{y}$) indicates prediction.

Each of the losses acting on valence and arousal are further decomposed as follows:
\begin{equation*}
    \mathcal{L}_{MSE}(\boldsymbol{y}_{C}, \hat{\boldsymbol{y}}_{C}) = MSE(\boldsymbol{y}_{Val}, \hat{\boldsymbol{y}}_{Val}) + MSE(\boldsymbol{y}_{Ar}, \hat{\boldsymbol{y}}_{Ar})
\end{equation*}
\begin{equation*}
    \mathcal{L}_{PCC}(\boldsymbol{y}_{C}, \hat{\boldsymbol{y}}_{C}) = 1 - \frac{PCC(\boldsymbol{y}_{Val}, \hat{\boldsymbol{y}}_{Val}) + PCC(\boldsymbol{y}_{Ar}, \hat{\boldsymbol{y}}_{Ar})}{2}
\end{equation*}
\begin{equation*}
    \mathcal{L}_{CCC}(\boldsymbol{y}_{C}, \hat{\boldsymbol{y}}_{C}) = 1 - \frac{CCC(\boldsymbol{y}_{Val}, \hat{\boldsymbol{y}}_{Val}) + CCC(\boldsymbol{y}_{Ar}, \hat{\boldsymbol{y}}_{Ar})}{2}
\end{equation*}





%% file: sections/results.tex
\section{RESULTS}
In all tables, for Arousal and Valence we report the CCC, and for Accuracy the portion of correctly classified examples on the discrete apparent emotional facial expressions.

\vspace{-5pt}
\subsection{Model Selection}

\begin{figure}
    \centering
    \includegraphics[scale=0.575]{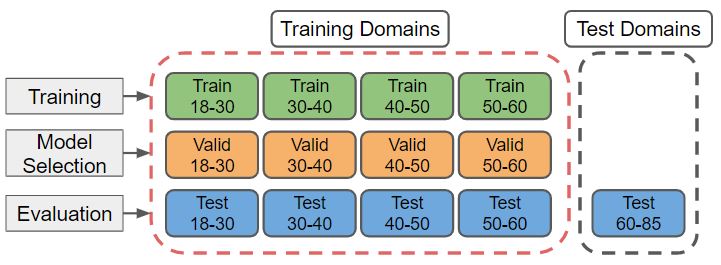}
    \caption{The algorithms train using data from the training-sets of the training domains. Model selection is then performed using the validation-sets of the training domains. Only at test time the algorithms observe data coming from the test-domain [(60-85)], where we evaluate performance on (Test 60-85).}
    \label{fig:dg_explain}
\end{figure}

In this work model-selection is carried out based on the performance of the model on the validation-sets of the training domains. In Figure \ref{fig:dg_explain} we present an example for clarity.




In our experiments the total size of the training-sets remains the same. For example, in two scenarios, in the first we train on [(18-30), (30-40), (40-50), (50-60)] and in the second we train on [(18-30), (30-40), (40-50)]. Then the cumulative size of training-sets in the first case is the same as the cumulative size of training-sets from the second. 


\subsection{Experiments}
\subsubsection{Leave-One-Domain-Out Algorithm Comparison}\label{subsection:exp_lodo_comparison}\hfill

In Table \ref{tab:leave_one_domain_out} we compare the algorithms on a leave-one-domain-out basis. For these experiments we considered five different settings, each with one of the age-groups being absent from training. The first column of Table \ref{tab:leave_one_domain_out} refers to training \textbf{without} age-group [18-30], and the reported values correspond to the test-set of age-group [18-30]. 
CDANN comes out on top four out of five configurations, and Mixup once. One possible explanation is that CDANN makes use of more information - the remaining DG algorithms make use of the features, while CDANN makes use of the discrete apparent facial emotion classes as well.

\begin{table}[!t]
    \centering
    \begin{tabular}{l|c|@{\hspace{1\tabcolsep}}c@{\hspace{1\tabcolsep}}ccc@{\hspace{1\tabcolsep}}c|@{\hspace{1\tabcolsep}}c|@{\hspace{1\tabcolsep}}}
\hline
\hline
 & \specialcell{\textbf{Eval.}\\\textbf{on}~$\rightarrow$} &  \textbf{18-30} &  \textbf{30-40} &  \textbf{40-50} &  \textbf{50-60} &  \textbf{60-85} & \textbf{Mean}\\
\hline
\hline
\textbf{CDANN} & \textbf{Ar}  & \cellcolor{red!10}0.61 & \cellcolor{red!10}0.64 & 0.55 & \cellcolor{red!10}0.51 & \cellcolor{red!10}0.57 &  0.58 \\
      & \textbf{Val} & \cellcolor{red!10}0.71 & \cellcolor{red!10}0.67 & 0.62 & \cellcolor{red!10}0.61 & \cellcolor{red!10}0.64 &  0.65 \\
      & \textbf{Acc} & \cellcolor{red!10}0.60 & \cellcolor{red!10}0.59 & 0.54 & \cellcolor{red!10}0.57 & \cellcolor{red!10}0.60 &  0.58 \\\hline
      
\textbf{ERM}   & \textbf{Ar}  & 0.54 & 0.61 &  0.59 &                    0.44 & 0.46 &  0.53 \\
      & \textbf{Val} & 0.69 & 0.63 &  0.62 &                    0.65 & 0.62 &  0.64 \\
      & \textbf{Acc} & 0.55 & 0.53 &  0.57 &                    0.56 & 0.53 &  0.55 \\\hline
      
\textbf{MMD}   & \textbf{Ar}  & 0.60 & 0.55 &  0.53 &                    0.48 & 0.50 &  0.53 \\
      & \textbf{Val} & 0.68 & 0.61 &  0.63 &                    0.61 & 0.62 &  0.63 \\
      & \textbf{Acc} & 0.57 & 0.53 &  0.57 &                    0.52 & 0.56 &  0.55 \\\hline
      
\textbf{MLDG}  & \textbf{Ar}  & 0.59 & 0.63 &  0.56 &                    0.49 & 0.54 &  0.56 \\
      & \textbf{Val} & 0.67 & 0.61 &  0.63 &                    0.59 & 0.66 &  0.63 \\
      & \textbf{Acc} & 0.55 & 0.54 &  0.57 &                    0.52 & 0.60 &  0.56 \\\hline
      
\textbf{Mixup} & \textbf{Ar}  & 0.63 & 0.60 &  \cellcolor{red!10}0.57 &  0.46 & 0.51 &  0.55 \\
      & \textbf{Val} & 0.71 & 0.65 &  \cellcolor{red!10}0.64 &  0.61 & 0.64 &  0.65 \\
      & \textbf{Acc} & 0.58 & 0.58 &  \cellcolor{red!10}0.56 &  0.57 & 0.57 &  0.57 \\\hline
\end{tabular}
    \caption{We compare the algorithms on a leave-one-domain-out basis. \textbf{Every column corresponds to that particular age-group being left out of training, and then evaluating on that age-group's test-set.} The train-sets of the four age-groups were used for training, then their validation-sets were used for model selection, and then we compare performance on the test-set of the unseen age-group. Colour indicates the best performing algorithm per setting. For example, in the first column the algorithms trained \textbf{without} age-group [18-30], and then evaluated on the test-set of age-group [18-30].
    }
    \label{tab:leave_one_domain_out}
\vspace{-20pt}
\end{table}

\subsubsection{Leave-One-Domain-Out Generalisation}\label{subsection:exp_lodo_gen}\hfill

In the experiments that follow we focus only on CDANN since it is the best performing algorithm. In Table \ref{tab:comparing_four_to_five} we present more detailed results related to leave-one-domain-out experiments. As expected when an age-group is absent from training the performance on its test-set drops (with the exception of the [60-85] age-group).

\begin{table}[!t]
    \centering
    \begin{tabular}{c|c|ccccc}
\hline
\hline
\specialcell{\textbf{Training}\\\textbf{Domains}~$\downarrow$} & \specialcell{\textbf{Evaluating}\\\textbf{on}~$\rightarrow$} &  \textbf{18-30} &  \textbf{30-40} &  \textbf{40-50} &  \textbf{50-60} &  \textbf{60-85} \\
\hline
\hline
\textbf{Without}    & \textbf{Arousal}   &         0.66 &         0.68 &         0.59 &         0.50 &         \cellcolor{violet!10}0.57 \\
\textbf{60-85}      & \textbf{Valence}  &         0.74 &         0.70 &         0.66 &         0.63 &        \cellcolor{violet!10}0.64 \\
                    & \textbf{Accuracy}  &         0.65 &         0.63 &         0.61 &         0.57 &         \cellcolor{violet!10}0.60 \\\cline{2-7}
                    & \textbf{Loss} &         1.32 &         1.32 &         1.47 &         1.66 &         1.51 \\\hline
           
\textbf{Without}    & \textbf{Arousal}   &     0.67 &         0.66 &         0.59 &         \cellcolor{orange!10}0.51 &         0.48 \\
\textbf{50-60}      & \textbf{Valence}  &      0.74 &         0.71 &         0.65 &         \cellcolor{orange!10}0.61 &         0.58 \\
                    & \textbf{Accuracy}  &      0.66 &         0.62 &         0.60 &         \cellcolor{orange!10}0.57 &         0.56 \\\cline{2-7}
                    & \textbf{Loss} &       1.27 &         1.35 &         1.55 &         1.69 &         1.78 \\\hline
           
\textbf{Without}    & \textbf{Arousal}   &       0.64 &         0.64 &         \cellcolor{green!10}0.55 &         0.48 &         0.51 \\
\textbf{40-50}      & \textbf{Valence}  &        0.72 &         0.67 &         \cellcolor{green!10}0.62 &         0.61 &         0.63 \\
                    & \textbf{Accuracy}  &        0.59 &         0.59 &         \cellcolor{green!10}0.54 &         0.53 &         0.56 \\\cline{2-7}
                    & \textbf{Loss} &         1.46 &         1.50 &         1.71 &         1.80 &         1.70 \\\hline
           
\textbf{Without}    & \textbf{Arousal}   &  0.63 & \cellcolor{red!10}0.64 & 0.56 & 0.50 & 0.48 \\
\textbf{30-40}      & \textbf{Valence}  & 0.72 &  \cellcolor{red!10}0.67 & 0.61 & 0.61 &  0.59 \\
                    & \textbf{Accuracy}  & 0.60 & \cellcolor{red!10}0.59 & 0.58 & 0.56 & 0.58 \\\cline{2-7}
                    & \textbf{Loss} & 1.48 & 1.56 & 1.71 & 1.72 & 1.82 \\\hline
    
\textbf{Without}    & \textbf{Arousal}   &         \cellcolor{blue!10}0.61 &         0.63 &         0.56 &         0.48 &         0.48 \\
\textbf{18-30}      & \textbf{Valence}  &         \cellcolor{blue!10}0.71 &         0.67 &         0.64 &         0.60 &        0.60 \\
                    & \textbf{Accuracy}  &         \cellcolor{blue!10}0.60 &         0.61 &         0.58 &         0.53 &         0.57 \\\cline{2-7}
                    & \textbf{Loss} &         1.55 &         1.54 &         1.69 &         1.83 &         1.83 \\\hline\hline
    
\textbf{With All}   & \textbf{Arousal}            &         \cellcolor{blue!10}0.67 &         \cellcolor{red!10}0.67 &         \cellcolor{green!10}0.58 &         \cellcolor{orange!10}0.51 &         \cellcolor{violet!10}0.52 \\
    & \textbf{Valence} &         \cellcolor{blue!10}0.74 &         \cellcolor{red!10}0.71 &         \cellcolor{green!10}0.66 &         \cellcolor{orange!10}0.64 &         \cellcolor{violet!10}0.61 \\
    & \textbf{Accuracy} &         \cellcolor{blue!10}0.65 &         \cellcolor{red!10}0.63 &         \cellcolor{green!10}0.60 &         \cellcolor{orange!10}0.56 &         \cellcolor{violet!10}0.62 \\\cline{2-7}
        & \textbf{Loss} &         1.29 &         1.31 &         1.50 &         1.60 &         1.59 \\\hline
\end{tabular}
    \caption{Experiments comparing training with four age-groups against training with all five age-groups. \textbf{Reported values correspond to each age-group's (column name) test-set.} Top five sets of rows correspond to training without the specified age-group, while the bottom set corresponds to training with all five. \textbf{Colours allow for quick identification of performance on a age-group's test-set, when the age-group is absent from training.}
    }
    \label{tab:comparing_four_to_five}
\vspace{-10pt}
\end{table}
Table \ref{tab:loss_difference} shows the \textit{decrease} in performance across the five age-groups, when each one of them (in turn) is left out of the training. This allows us to see the relative importance of each of the age-groups for the performance on the remaining age-groups. We observe that excluding a age-group from training tends have a greater impact on its nearest (age-wise) age-groups (With the exception of [60-85]), probably due to a higher level of similarity between facial features and a more similar way of apparent emotional facial expression recognition. For both scenarios we see that age-group [60-85] is anomalous. One plausible explanation for this is the sizes of the validation and test-sets for this particular age-group which amount to $160$ and $169$ respectively, are small and therefore results could be misleading.

\begin{table}[!t]
    \centering
    \begin{tabular}{cc|ccccc}
    \hline
 \specialcell{Training\\Domains~$\downarrow$} & \specialcell{Eval.\\on~$\rightarrow$} & \textbf{18-30} &  \textbf{30-40} &  \textbf{40-50} &  \textbf{50-60} &  \textbf{60-85} \\
    \hline\hline
    Without [18-30] & +Loss & \cellcolor{blue!10}0.26 & \cellcolor{blue!10}0.23 & \cellcolor{blue!10}0.19 & \cellcolor{blue!10}0.22 & \cellcolor{blue!10}0.25 \\
    Without [30-40] & +Loss & \cellcolor{red!10}0.18 & \cellcolor{red!10}0.20 & \cellcolor{red!10}0.16 & \cellcolor{red!10}0.14 & \cellcolor{red!10}0.17 \\
    Without [40-50] & +Loss & \cellcolor{green!10}0.17 & \cellcolor{green!10}0.19 & \cellcolor{green!10}0.20 & \cellcolor{green!10}0.20 & \cellcolor{green!10}0.12 \\
    Without [50-60] & +Loss & \cellcolor{orange!10}-0.02 & \cellcolor{orange!10}0.04 & \cellcolor{orange!10}0.05 & \cellcolor{orange!10}0.08 & \cellcolor{orange!10}0.20 \\
    Without [60-85] & +Loss & \cellcolor{violet!10}0.03 & \cellcolor{violet!10}0.01 & \cellcolor{violet!10}-0.03 & \cellcolor{violet!10}0.06 & \cellcolor{violet!10}-0.07 \\
    \hline
    \end{tabular}
    \caption{Shows the \textbf{decrease} in performance (increase in $Loss$) when each of the age-groups is left out from training. For example, the first row of this table corresponds to subtracting the rows corresponding to the Loss for [Without 18-30] and [All] from Table \ref{tab:comparing_four_to_five}.
    }
    \label{tab:loss_difference}
\vspace{-10pt}
\end{table}

\begin{table}[!h]
    \centering
    \begin{tabular}{c|c@{\hspace{1.15\tabcolsep}}c@{\hspace{1.15\tabcolsep}}c@{\hspace{1.15\tabcolsep}}c@{\hspace{1.15\tabcolsep}}c@{\hspace{1.15\tabcolsep}}|c}
    \hline
    {} &  \textbf{18-30} &  \textbf{30-40} &  \textbf{40-50} &  \textbf{50-60} &  \textbf{60-85} &    \textbf{Total} \\
    \hhline{=======}
    \textbf{Without} \textbf{[50-60]} &  61.9 &  50.4 &  37.5 &      0 &  20.6 &  170.4 \\\hline
    \textbf{Without} \textbf{[40-50]} &  68.5 &  55.8 &      0 &  26.0 &  20.6 &  170.8 \\\hline
    \textbf{Without} \textbf{[40-60]} &  83.6 &  67.8 &      0 &      0 &  20.6 &  171.9 \\\hline
    \end{tabular}
    \caption{Training-set sizes (number of images in 1000s) for these configurations.}
    \label{tab:example_training_sizes}
\vspace{-20pt}
\end{table}

\subsubsection{Impact of training domains on unseen-domain-generalisation}\label{subsection:exp_unseen_domain}\hfill

In the results that follow we explore the impact the variety of age-groups during training has on out-of-domain-generalisation performance. We now focus on the change of performance when moving from three training age-groups, to four training age-groups. In Table \ref{tab:example_training_sizes} we show the sizes of the training-sets used in each of these three configurations.

We show results for four such configurations in Tables \ref{tab:1_2_112}, \ref{tab:1_3_113}, \ref{tab:0_3_103} \& \ref{tab:0_2_102}. An example on how to read the tables: for Table \ref{tab:1_2_112} column [50-60] we see the performance metrics for the test-set of age-group [50-60]. The two triplets that are coloured pink have in common that age-group [50-60] was not used during training, or validation. The comparison of these two triplets tells us about the effect of including age-group [40-50] on the out-of-domain-generalisation performance on age-group [50-60]. 

We compare performance between training with three age-groups and training with four age-groups. We see that in seven out of eight such comparisons, we observe an increase in performance on the unseen age-group, which provides evidence of impact of the number of training domains on out-of-domain-generalisation. This is probably an indication that the model extracts features that are `invariant' to specific age-groups, the more domains it sees.
\begin{table}[!h]
    \centering
    \begin{tabular}{c@{\hspace{1.5\tabcolsep}}|c@{\hspace{1.5\tabcolsep}}|c@{\hspace{1.5\tabcolsep}}c@{\hspace{1.5\tabcolsep}}cc@{\hspace{1.5\tabcolsep}}c}
\hline
\hline
\specialcell{\textbf{Training}\\\textbf{Domains}~$\downarrow$} & \specialcell{\textbf{Eval.}\\\textbf{on}~$\rightarrow$} &  \textbf{18-30} &  \textbf{30-40} &  \textbf{40-50} &  \textbf{50-60} &  \textbf{60-85} \\
\hline
\hline
\textbf{Without}    & Arousal  & 0.67 & 0.66 & 0.59 & \cellcolor{red!10}0.51 & 0.48 \\
\textbf{50-60}      & Valence & 0.74 & 0.71 & 0.65 & \cellcolor{red!10}0.61 & 0.58 \\
             & Accuracy & 0.66 & 0.62 & 0.60 & \cellcolor{red!10}0.57 & 0.56 \\\hline
                    
\textbf{Without}    & Arousal  & 0.64 & 0.64 & \cellcolor{blue!10}0.55 & 0.48 & 0.51 \\
\textbf{40-50}      & Valence & 0.72 & 0.67 & \cellcolor{blue!10}0.62 & 0.61 & 0.63 \\
             & Accuracy & 0.59 & 0.59 & \cellcolor{blue!10}0.54 & 0.53 & 0.56 \\\hline\hline
                    
\textbf{Without}    & Arousal  & 0.64 & 0.65 & \cellcolor{blue!10}0.53 & \cellcolor{red!10}0.49 & 0.44 \\
\textbf{40-50}      & Valence & 0.74 & 0.68 & \cellcolor{blue!10}0.60 & \cellcolor{red!10}0.61 & 0.59 \\
\textbf{\&50-60}    & Accuracy & 0.60 & 0.61 & \cellcolor{blue!10}0.55 & \cellcolor{red!10}0.52 & 0.54 \\\hline
\end{tabular}
    \caption{Out-of-domain-generalisation results for age-groups [40-50] and [50-60]. Evaluation is on the test-set of each age-group.}
    \label{tab:1_2_112}
\vspace{-20pt}
\end{table}

\begin{table}[!h]
    \centering
    \begin{tabular}{c@{\hspace{1.5\tabcolsep}}|c@{\hspace{1.5\tabcolsep}}|c@{\hspace{1.5\tabcolsep}}c@{\hspace{1.5\tabcolsep}}c@{\hspace{1.5\tabcolsep}}c@{\hspace{1.5\tabcolsep}}c}
\hline
\hline
\specialcell{\textbf{Training}\\\textbf{Domains}~$\downarrow$} & \specialcell{\textbf{Eval.}\\\textbf{on}~$\rightarrow$} &  \textbf{18-30} &  \textbf{30-40} &  \textbf{40-50} &  \textbf{50-60} &  \textbf{60-85} \\
\hline
\hline
\textbf{Without}        & Arousal  & 0.67 & 0.66 & 0.59 & \cellcolor{red!10}0.51 & 0.48 \\
\textbf{50-60} & Valence & 0.74 & 0.71 & 0.65 & \cellcolor{red!10}0.61 & 0.58 \\
               & Accuracy & 0.66 & 0.62 & 0.60 & \cellcolor{red!10}0.57 & 0.56 \\\hline
                
\textbf{Without}        & Arousal  & 0.63 & \cellcolor{blue!10}0.64 & 0.56 & 0.50 & 0.48 \\
\textbf{30-40} & Valence & 0.72 & \cellcolor{blue!10}0.67 & 0.61 & 0.61 & 0.59 \\
               & Accuracy & 0.60 & \cellcolor{blue!10}0.59 & 0.58 & 0.56 & 0.58 \\\hline\hline
               
\textbf{Without}          & Arousal  & 0.63 & \cellcolor{blue!10}0.60 & 0.52 & \cellcolor{red!10}0.44 & 0.47 \\
\textbf{30-40}   & Valence & 0.72 & \cellcolor{blue!10}0.66 & 0.62 & \cellcolor{red!10}0.59 & 0.60 \\
\textbf{\&50-60} & Accuracy & 0.62 & \cellcolor{blue!10}0.57 & 0.56 & \cellcolor{red!10}0.56 & 0.56 \\\hline

\end{tabular}
    \caption{Out-of-domain-generalisation results for age-groups [30-40] and [50-60]. Evaluation is carried on the test-set of each age-group.}
    \label{tab:1_3_113}
\vspace{-20pt}
\end{table}

\begin{table}[!h]
    \centering
    \begin{tabular}{c@{\hspace{1.5\tabcolsep}}|c@{\hspace{1.5\tabcolsep}}|c@{\hspace{1.5\tabcolsep}}c@{\hspace{1.5\tabcolsep}}c@{\hspace{1.5\tabcolsep}}c@{\hspace{1.5\tabcolsep}}c}
\hline
\hline
\specialcell{\textbf{Training}\\\textbf{Domains}~$\downarrow$} & \specialcell{\textbf{Eval.}\\\textbf{on}~$\rightarrow$} &  \textbf{18-30} &  \textbf{30-40} &  \textbf{40-50} &  \textbf{50-60} &  \textbf{60-85} \\
\hline
\hline
\textbf{Without}             & Arousal  & 0.66 & 0.68 & 0.59 & 0.50 & \cellcolor{red!10}0.57 \\
\textrm{60-85}      & Valence & 0.74 & 0.70 & 0.66 & 0.63 & \cellcolor{red!10}0.64 \\
                    & Accuracy & 0.65 & 0.63 & 0.61 & 0.57 & \cellcolor{red!10}0.60 \\\hline
                    
\textbf{Without}             & Arousal  & 0.63 & \cellcolor{blue!10}0.64 & 0.56 & 0.50 & 0.48 \\
\textbf{30-40}      & Valence & 0.72 & \cellcolor{blue!10}0.67 & 0.61 & 0.61 & 0.59 \\
                    & Accuracy & 0.60 & \cellcolor{blue!10}0.59 & 0.58 & 0.56 & 0.58 \\\hline\hline
                    
\textbf{Without}             & Arousal  & 0.65 & \cellcolor{blue!10}0.64 & 0.56 & 0.51 & \cellcolor{red!10}0.47 \\
\textbf{30-40}      & Valence & 0.71 & \cellcolor{blue!10}0.65 & 0.62 & 0.61 & \cellcolor{red!10}0.59 \\
\textbf{\&60-85}    & Accuracy & 0.59 & \cellcolor{blue!10}0.58 & 0.55 & 0.53 & \cellcolor{red!10}0.53 \\\hline
\hline
\end{tabular}
    \caption{Out-of-domain-generalisation results for age-groups [30-40] and [60-85]. Evaluation is on the test-set of each age-group.}
    \label{tab:0_3_103}
\vspace{-20pt}
\end{table}

\begin{table}[!h]
    \centering
    \begin{tabular}{c@{\hspace{1.5\tabcolsep}}|c@{\hspace{1.5\tabcolsep}}|c@{\hspace{1.5\tabcolsep}}c@{\hspace{1.5\tabcolsep}}c@{\hspace{1.5\tabcolsep}}c@{\hspace{1.5\tabcolsep}}c}
\hline
\hline
\specialcell{\textbf{Training}\\\textbf{Domains}~$\downarrow$} & \specialcell{\textbf{Eval.}\\\textbf{on}~$\rightarrow$} &  \textbf{18-30} &  \textbf{30-40} &  \textbf{40-50} &  \textbf{50-60} &  \textbf{60-85} \\
\hline
\hline
\textbf{Without}             & Arousal  & 0.66 & 0.68 & 0.59 & 0.50 & \cellcolor{red!10}0.57 \\
\textbf{60-85}      & Valence & 0.74 & 0.70 & 0.66 & 0.63 & \cellcolor{red!10}0.64 \\
                    & Accuracy & 0.65 & 0.63 & 0.61 & 0.57 & \cellcolor{red!10}0.60 \\\hline
                    
\textbf{Without}             & Arousal  & 0.64 & 0.64 & \cellcolor{blue!10}0.55 & 0.48 & 0.51 \\
\textbf{40-50}      & Valence & 0.72 & 0.67 & \cellcolor{blue!10}0.62 & 0.61 & 0.63 \\
                    & Accuracy & 0.59 & 0.59 & \cellcolor{blue!10}0.54 & 0.53 & 0.56 \\\hline\hline
                    
\textbf{Without}             & Arousal  & 0.64 & 0.66 & \cellcolor{blue!10}0.56 & 0.50 & \cellcolor{red!10}0.41 \\
\textbf{40-50}      & Valence & 0.71 & 0.69 & \cellcolor{blue!10}0.63 & 0.64 & \cellcolor{red!10}0.62 \\
\textbf{\&60-85}    & Accuracy & 0.62 & 0.61 & \cellcolor{blue!10}0.54 & 0.56 & \cellcolor{red!10}0.50 \\\hline

\hline
\end{tabular}
    \caption{Out-of-domain-generalisation results for age-groups [40-50] and [60-85]. Evaluation is on the test-set of each age-group.}
    \label{tab:0_2_102}
\vspace{-20pt}
\end{table}

%% file: sections/conclusion.tex
\section{CONCLUSION}
In this work we study the task of domain generalisation in the context of apparent emotional facial expression recognition amongst different age groups. We first compare domain generalisation algorithms on the grounds of out-of-domain-generalisation. We then examine the effect the variety and number of training age-groups has on the performance on unseen age-groups. The results provide evidence that an increase in the number of training age-groups does indeed help with apparent emotional facial expression recognition performance on unseen age-groups. Through comparing different configurations of the training age-group we show that exclusion of an age group tends to affect more the apparent emotional facial expression recognition performance of the neighbouring age groups.